\documentclass[conference]{IEEEtran}
\IEEEoverridecommandlockouts
\usepackage{cite}
\usepackage{amsmath,amssymb,amsfonts}
\usepackage{algorithmic}
\usepackage[table,xcdraw]{xcolor}

\usepackage{graphicx}
\usepackage{lipsum}
\usepackage{textcomp}
\ifCLASSOPTIONcompsoc
    \usepackage[caption=false, font=normalsize, labelfont=sf, textfont=sf]{subfig}
\else
\usepackage[caption=false, font=footnotesize]{subfig}
\fi
\usepackage{xcolor}
\def\BibTeX{{\rm B\kern-.05em{\sc i\kern-.025em b}\kern-.08em
    T\kern-.1667em\lower.7ex\hbox{E}\kern-.125emX}}
\begin{document}

\title{Naturalizing Neuromorphic Vision Event Streams Using GANs\\
}

\author{Dennis E. Robey$^{1}$, Wesley Thio$^{2}$, Herbert H.C. Iu$^{1}$, Jason~K.~Eshraghian$^{2}$\\

\IEEEauthorblockA{$^1$\textit{School of Electrical, Electronic and Computer Engineering, University of Western Australia, Perth, WA 6009 Australia}}
\IEEEauthorblockA{$^2$\textit{School of Electrical, Electronic and Computer Engineering, University of Michigan, Ann Arbor, MI 48109 USA}}}%

\maketitle

\begin{abstract}
Dynamic vision sensors are able to operate at high temporal resolutions within resource constrained environments, though at the expense of capturing static content. The sparse nature of event streams enables efficient downstream processing tasks as they are suited for power-efficient spiking neural networks. One of the challenges associated with neuromorphic vision is the lack of interpretability of event streams. While most application use-cases do not intend for the event stream to be visually interpreted by anything other than a classification network, there is a lost opportunity to integrating these sensors in spaces that conventional high-speed CMOS sensors cannot go. For example, biologically invasive sensors such as endoscopes must fit within stringent power budgets, which do not allow MHz-speeds of image integration. While dynamic vision sensing can fill this void, the interpretation challenge remains and will degrade confidence in clinical diagnostics. The use of generative adversarial networks presents a possible solution to overcoming and compensating for a vision chip's poor spatial resolution and lack of interpretability. In this paper, we methodically apply the Pix2Pix network to naturalize the event stream from spike-converted CIFAR-10 and Linnaeus 5 datasets. The quality of the network is benchmarked by performing image classification of naturalized event streams, which converges to within 2.81\% of equivalent raw images, and an associated improvement over unprocessed event streams by 13.19\% for the CIFAR-10 and Linnaeus 5 datasets.

\end{abstract}

\section{Introduction}
Neuromorphic vision seeks to draw principles from the biological retina to improve the efficiency with which we capture images. The retina integrates over 100 million photoreceptor cells, operating over a dynamic range of 180~dB within a stringent thermal limit of 3~mW \cite{Baek2020AICAS, nonlinear, formulation, arrow2020}. There is no digital camera that comes close to these metrics. But the onset of the dynamic vision sensor (DVS) has proven to be one possible answer in achieving the optimal silicon retina \cite{Lichsteiner2008, Brandli2014, Yang2015, Eshraghian2018, Azghadi2020}. Dynamic vision sensors operate on the hypothesis that the biological retina only processes images when there is a change of input. In this way, it manages to limit its power consumption. In contrast, a CMOS image sensor will oversample unchanging data and undersample fast-moving content by allocating equivalent resources to each pixel and synchronously processing them all without discrimination of the scene being captured. The data transmission rate of CMOS image sensors is often wasted on unchanging input.

Event-driven sensing removes waste by using address event representation to asynchronously process a changing pixel independent of any frame. By removing redundant data, this promotes better allocation of resources to achieving high temporal resolution with less power dissipation. However, this optimization comes with the cost of reduced interpetability of images. While this is not necessarily an issue in many applications, there are numerous domains that would benefit from having interpretable image sequences of high temporal resolution. The most obvious example is for invasive biomedical sensing. Micro cameras in endoscopies must operate within strict power budgets to mitigate potential harm to the patient. The microscale form factor and prohibitive power budget imposes a significant limitation on the maximum spatiotemporal resolution. Practically, this is burdensome for surgeons as higher spatial resolution results in increased depth perception and better hand-eye coordination. Faster temporal resolution opens up the possibility to capture biomarkers that have not thus far been possible. 

While neuromorphic sensing can fill this void, the most obvious challenge is that event streams are difficult to interpret and classify. This makes them prohibitive in many critical applications, such as a minimally invasive surgical camera. Naturalizing event streams using generative adversarial networks (GANs) presents an opportunity to restore binary events to what a CMOS image sensor would capture, but with the use of a low-power and high-speed DVS \cite{EshraghianMI, EshraghianMI2}. Here, we methodically apply the Pix2Pix algorithm to naturalize event streams from spiking CIFAR-10 and Linnaeus 5 datasets, and benchmark performance of the GAN in order to quantify the potential improvement of classification accuracy. Each image is classified using a separately trained network which demonstrates an improvement over unprocessed event streams of 13.19\%, and converging within 2.81\% of non-spiking CMOS images. This demonstrates the promise of deploying DVS cameras beyond their intended application.

Section II provides the relevant background on GANs, followed by section III which describes the methods employed to conditionally generate images from spikes, as well as the approach taken to perform classification. Finally, our experimental results are provided which provably shows the potential of merging GANs with event streams. We note that we will use the terms `spikes' and `events' interchangeably to refer to discrete binary instances of activity. 

\begin{figure*}[!ht]
\centering
\includegraphics[scale=0.45]{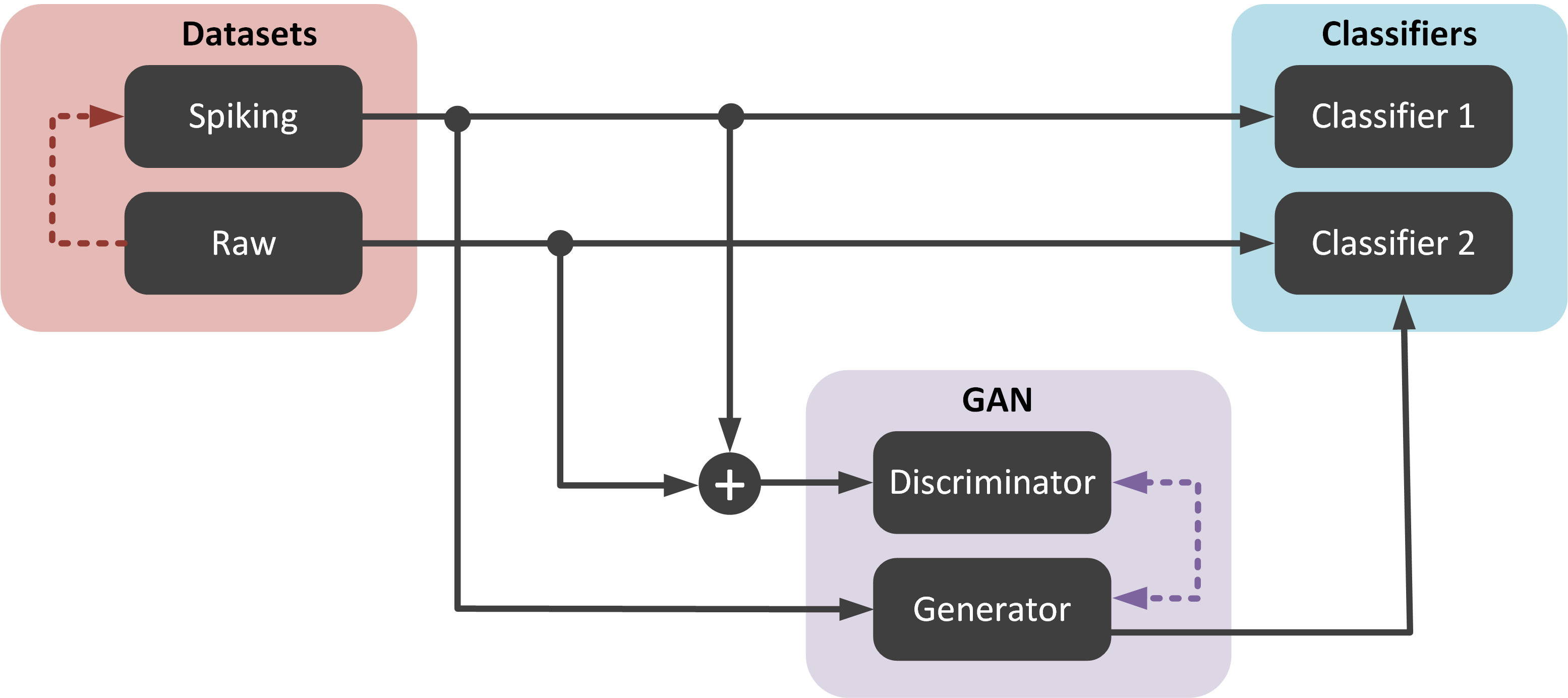}
\caption{Raw and spiking datasets are both benchmarked by finding their test set accuracy prior to synthesizing the raw data from the spiking data.}\label{fig:method}
\end{figure*}

\section{Generative Adversarial Networks}
The original formulation of GANs by Ian Goodfellow \textit{et al.} \cite{Goodfellow2014} implemented the idea of training two models against each other. The generative model \(G\) tries to capture the target data distribution whereas the discriminative model \(D\) estimates the likelihood that a sample came from training data rather than \(G\). In mathematical terms, \(D\) and \(G\) are in a two-player minimax game with value function \(V(D, G)\):

\begin{equation}
    \begin{split}
    min_G max_D V(D, G) = E_{x \sim p_{data}(x)} \left[ \log D(x) \right] \\
    + E_{z \sim p_z(z)} \left[ \log \left( 1 - D(G(z)) \right)\right],
    \end{split}
\label{eq:gan_loss_function}\end{equation}

\noindent where \(p_g\) is the generator's distribution over data \(x\), and \(p_z(z)\) is the input noise variable. In 2018, Phillip Isola \textit{et al.}~improved upon this idea by using a conditional GAN (cGAN) that learns the mapping between pairs of images \cite{Isola2016}. This GAN is \emph{conditional} because a conditional input image \(x\) is fed to both generator and discriminator, as shown in (2):

\begin{equation}
    \begin{split}
    \mathcal{L}_{GAN} (G,D) = \mathbb{E}_{x,y} [\log D(x,y)] \\
        +\mathbb{E}_{x,z} [\log (1 - D(x, G(x,z))], \\
    \end{split}
\label{eq:cgan_loss_function}\end{equation}

\noindent where \(x\) is a conditional source image, \(y\) is the ground truth target image, and \(z\) is the random noise vector. What makes Pix2Pix unique is the architecture; the generator is based on a \emph{U-Net} structure and the discriminator uses a convolutional \emph{PatchGAN} classifier \cite{Isola2018}.

\section{Methods}
Spiking datasets have been developed in the past by either filming static datasets using a moving DVS to emulate saccades \cite{Orchard2015} or by algorithmically generating events from a dataset, such as by passing pixel intensities into a binomial function to emulate a Poisson spike train. Here, we use 3 categories of spiking and emulated spiking datasets that are benchmarked against each other, in addition to the unaltered ground truth image and the regenerated full color image generated by the cGAN.
\begin{enumerate}
\item
  Greyscale images collected by processing event data procured from a
  vision chip. These images are created by integrating time-based spikes
  over a period to create a new sequence known as a \emph{time-surface}.
\item
  Black and white edges generated from a Canny edge detector algorithm, serving as a comparison to the edges that are generated from
  the event data.
\item
  Greyscale edges generated with Holistically-Nested Edge Detection
  (HED) that simulates time-surface generation.
\end{enumerate}

The high-level approach to benchmarking is shown in Fig.~\ref{fig:method}. The raw and spiking datasets will both be independently used to train a pair of classifiers. The discriminator of the GAN will use coupled pairs of raw and spiking data, with the event treated as the conditional source ($x$ in (2)), and the raw data is the ground truth target image ($y$ in (2)). The spiking dataset is characteristic of what would be captured by a DVS, and the raw data of that by a conventional CMOS image sensor. The generator is then used to produce samples which are classified by `Classifier 2', and are compared to the accuracy of equivalent samples from the raw dataset.

We use the CIFAR-10 and Linnaeus 5 \cite{Chaladze2017} datasets as the raw datasets. The use of both datasets increases our confidence in the generalization of our approach. These are preferable over MNIST for the richer features present in the images, and circumvents stability problems associated with training GANs on much higher-dimensional datasets such as ImageNet. The low resolution of both datasets is analogous to the low-resolution of most commercially available DVS \cite{Inivation2019}, and can be used to evaluate the upsampling capacity of the GAN. The CIFAR10-DVS dataset \cite{Li2017} was created by sampling images from the CIFAR-10 dataset, totalling 10,000 event streams split evenly across 10 classes. We have opted to test both DVS and algorithmic approaches to generating time surfaces so as to have variations in spiking data for the same raw input. We expect the algorithmic approach may reduce overfitting to specific samples by generating random samples for each epoch, although this remains to be seen \cite{Li2017}.

Altogether, the Linnaeus 5 dataset offers 12,000 images within 5 classes,
equalling 1,600 images per class. This division is modified for use when
training the GAN. In training a GAN, it is essential to train with objects that are
similar so that the neural network may have a chance to learn the
mapping from the source image to the target image, hence we cannot use
the ``other (miscellaneous)'' class. Furthermore, there should not be
several instances of the same object inside the same image. Birds and
dogs often have natural flora inside the images such as berries and
flowers hence the ``berry'' and ``flower'' classes are also omitted.
Thus the culled dataset only includes the ``bird'' and ``dog'' classes.
Furthermore, the data for each class is split into 800 training images, 400 validation images and 400 test images. 

Classification is based on the original 2014 VGG16 CNN architecture
\cite{Simonyan2014}. The final version optimized for the CIFAR-10 and spiking CIFAR-10 datasets had 6 convolutional layers, 3 pooling layers, 3 dropout layers, leaky ReLU activation, data augmentation and batch normalization. Generation of images is handled by the original Pix2Pix cGAN \cite{Isola2016, Zhu2017}. This in turn, consists of target ground truth images and source
edges that represent time-surfaces, generated via Canny edge detection and HED (via OpenCV library). Several CNNs are trained to classify different versions of the
Linnaeus 5 datasets: the original images, Canny, and HED edge detection. The classifier trained on the original images is then used to
classify the images generated from the cGAN. Fine tuning of hyperparameters is not performed so as to ensure a fair comparison across the various distributions of data on the same network.

\begin{figure}
\centering
\includegraphics[scale=0.5]{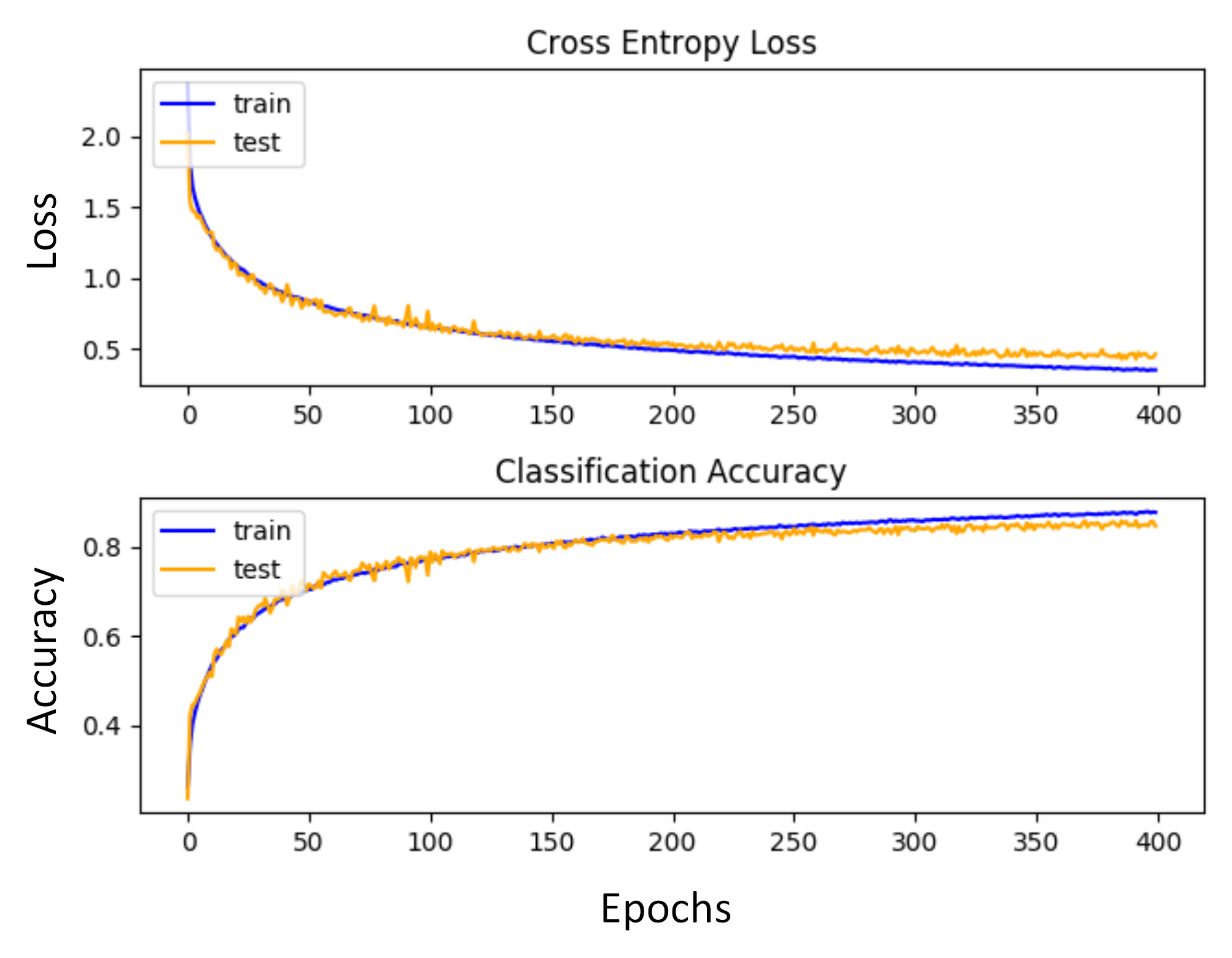}
\caption{Raw CIFAR-10 dataset loss and accuracy shows low variance. This same network is also used for inference only on synthesized images after training the GAN. These results will clearly be the optimal benchmark as the synthesized samples come from a different distribution.}\label{fig:cnn_raw}
\end{figure}

\section{Results}\label{results}



The training process of the raw CIFAR-10 dataset is shown in Fig.~\ref{fig:cnn_raw}, which shows a test set accuracy of 84.74\%. This serves as the upper-limit benchmark. Clearly, state-of-the-art is much higher than this, but as the synthesized data from the cGAN is sampled from a reconstructed distribution, it cannot be expected to perfectly match the raw inputs when used for inference on the same network.

\begin{figure}
\centering
\includegraphics[scale=0.5]{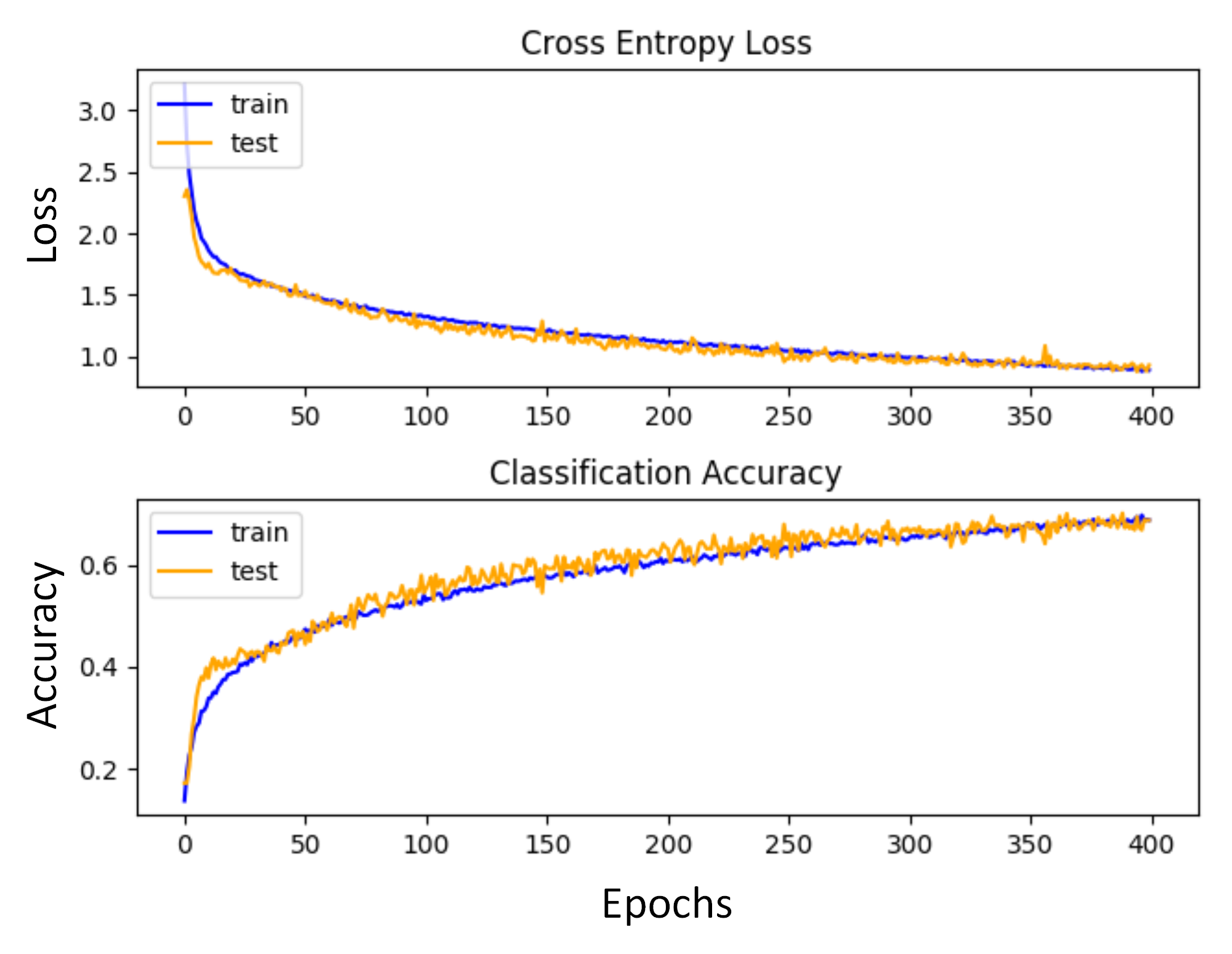}
\caption{CIFAR10-DVS training progress shows high bias due to lossy and sparse input data. The test set accuracy is less stable than that from the raw CIFAR-10 model.}\label{fig:cnn_dvs_results}
\end{figure}

Given that each sample from the DVS dataset contains lower spatial resolution and no color, unlike the raw input, we expect this to degrade test set accuracy.
The CIFAR10-DVS test set accuracy is 68.77\% on a separately trained network (with the same architecture as the raw case) and serves as the lower-limit benchmark, provided the cGAN is appropriately trained. The training process is shown in Fig.~\ref{fig:cnn_dvs_results}. Lower stability of loss is immediately visible, depicted by the jagged accuracy at test time. This can be attributed to the fact that the network is struggling to make accurate predictions due to the sparse nature of the data. High bias is expected to be a result of the lossy approach to spike conversion, and a lack of hyperparameter tuning. These are intentional to better emulate a real-world situation which has no ground truth to work with. We note that the state-of-the-art spiking neural network result on spike-converted CIFAR-10 is a classification error of 8.45\%, though this uses a near-lossless conversion technique as well as hyperparameter tuning \cite{Sengupta2019}. I.e., their goal is to reach state-of-the-art, whereas our goal is to closely match the test set accuracy of raw data as a pure result of using cGANs rather than by tuning, regularization and other optimization techniques. This setup better mimics the deployment of an image reconstruction system from events, where ground truths are not necessarily available to tune with.

\begin{figure}
\centering
\includegraphics[scale=0.4]{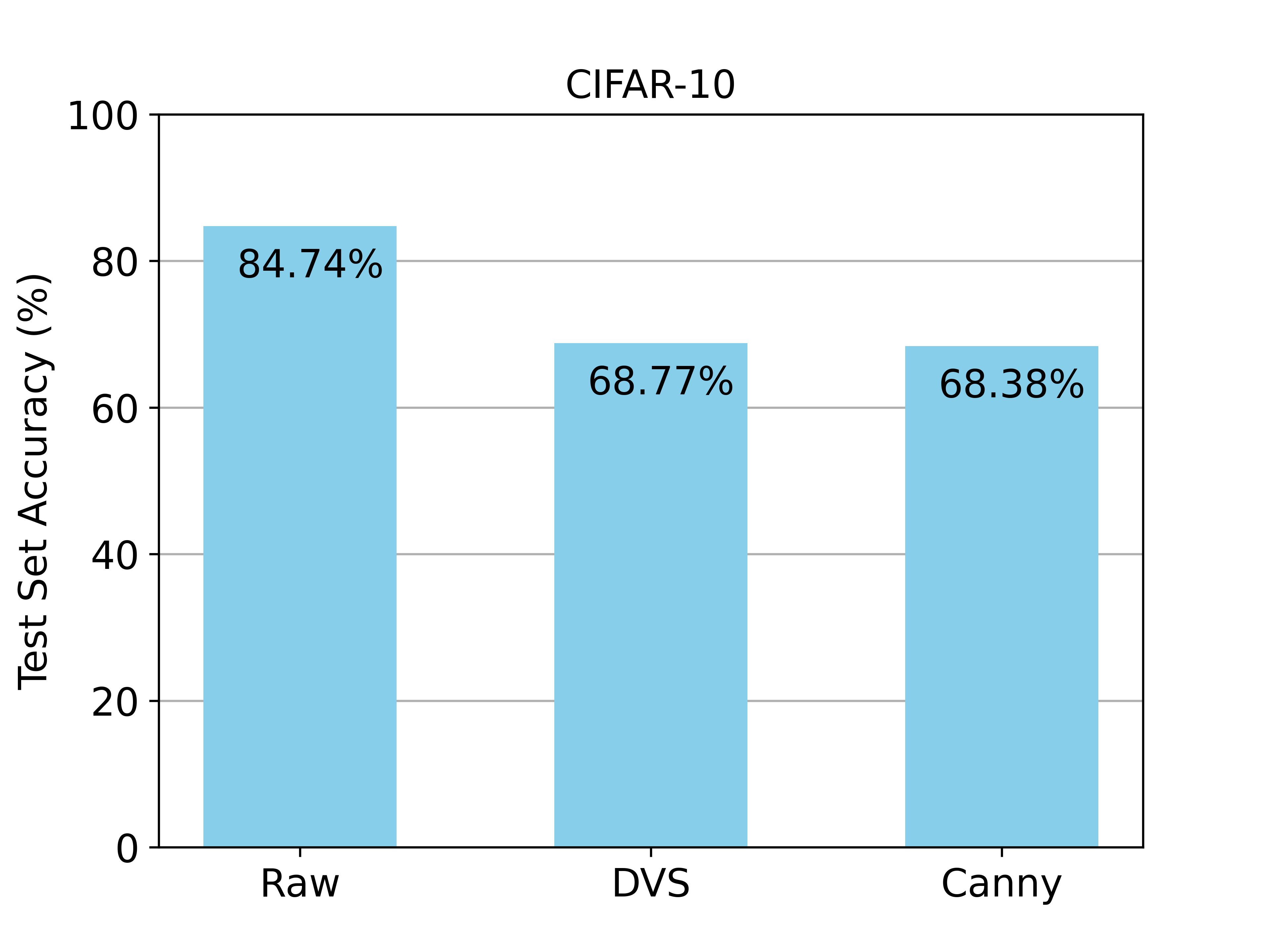}
\caption{CIFAR-10 classification accuracy on the raw input, DVS time-surfaces, and Canny spike-conversion approaches. The raw and DVS datasets were trained on their own networks, while Canny edge detection was tested on the DVS network showing low variance (though high bias). The latter two networks are used as a benchmark for cGAN synthesis.}\label{fig:plot_cifar10}
\end{figure}

Finally, for comparative purposes, Canny edge detected images were taken from the
CIFAR-10 dataset and classified with the spike-trained CNN. Despite being a completely different mode of spiking from a different distribution, the network appears to generalize much better than expected by generating results that are almost on 
par with the CIFAR10-DVS dataset, at a test set accuracy of 68.38\%. Our operating hypothesis for this is that each input feature may take on the value of an event or the absence of an event, thus restricting the domain each dimension of features may take on. 
Despite the high bias, this restriction appears to act in favor of reducing variance. The gap between the raw and spiking datasets is quite wide at approximately 16\% which gives us motivation to try and close this gap. The results are plotted in Fig.~\ref{fig:plot_cifar10}.

\begin{figure}[t]
\centering
\includegraphics[scale=0.675]{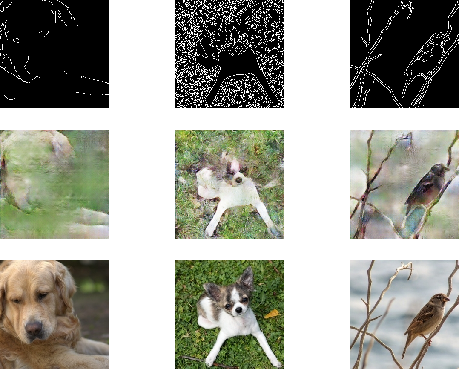}
\caption{Canny results of Pix2Pix. Top row: Canny edge detection. Middle row: Pix2Pix reconstruction. Bottom row: Ground truth/raw input. Note the poor performance in the left column due to lighting variation.}\label{fig:epoch_100}
\end{figure}

\begin{figure}[!t] 
    \centering
  \subfloat[\label{1a}]{%
       \includegraphics[width=0.22\linewidth]{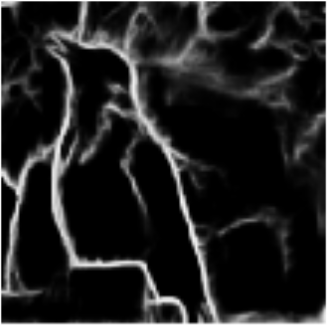}}
       \hfill
  \subfloat[\label{1b}]{%
        \includegraphics[width=0.22\linewidth]{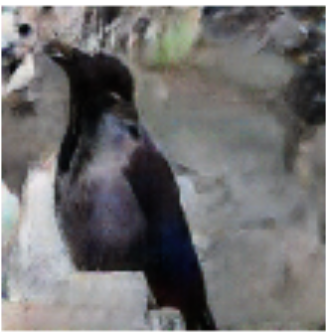}}
        \hfill
  \subfloat[\label{1c}]{%
        \includegraphics[width=0.22\linewidth]{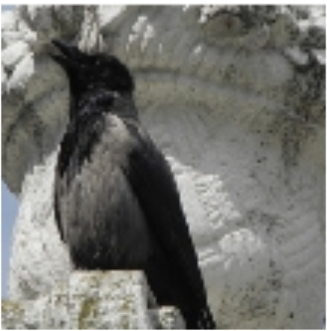}}
  \caption{HED results of Pix2Pix. (a) Spike synthesis using HED. (b) Ground truth. (c) Naturalized events.}
  \label{fig:epoch_hed} 
\end{figure}

An almost identical approach was taken for the Linnaeus 5 dataset, where Canny and HED edge detection were both fed into a cGAN and set to run for 200
epochs. Observe the three rows of images in Fig.~\ref{fig:epoch_100}.
The top row represents the source image, which in this case was the
Canny method. The bottom row represents the target ground truth image and
the middle row represents the image synthesized by the cGAN. One of the drawbacks of using the Canny edge detection algorithm
is that it does not perform well on images with varying lighting
conditions. The edges of the left side of the dog's face are
missing as the raw image is partially obscured by a shadow. This suggests that if the edge
data were better then the generated images would be better too. This is proven in Fig.~\ref{fig:epoch_hed} which uses the
same Pix2Pix GAN but with HED edges instead. Upon inspection, it can be seen
that the source image has less noise than Canny edge detection meaning the cGAN
has a better chance to map the image. As a result, the quality of the
synthesized image appears far closer to the target ground truth. The generated images for new datasets are compared with the
original.

\begin{figure}[!t]
\centering
\includegraphics[scale=0.4]{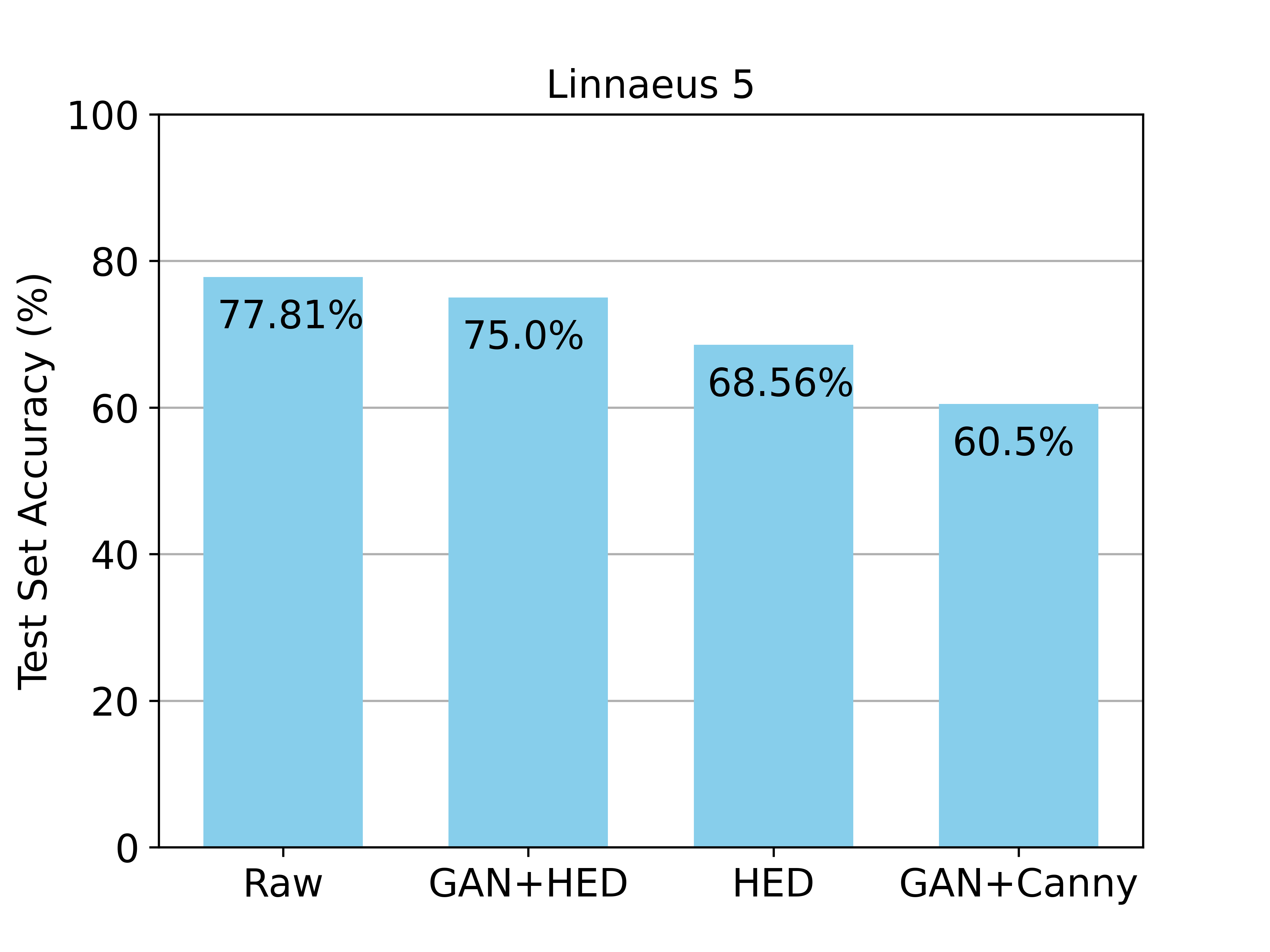}
\caption{Linnaeus 5 classification accuracy on the raw input, Pix2Pix reconstruction of HED events, HED events, and the reconstruction of Canny-derived events.}\label{fig:plot_linnaeus5}
\end{figure}

The final results are shown in Fig.~\ref{fig:plot_linnaeus5}. Note that
the result of using Pix2Pix with HED edges demonstrates performance on
par with that of raw images with merely 2.81\% difference
in accuracy without any hyperparameter tuning. However, it should be noted that the accuracy of
the Linnaeus 5 dataset overall is lower than that of CIFAR-10. This can
be attributed to the fact that the Linnaeus 5 dataset has considerably
smaller samples, especially when the extraneous classes were removed.
It should be noted too that the Pix2Pix reconstruction of Canny-based spikes performed the
worst. This is due to sparsity arising as a result of lighting variation noted before, which would not occur if recorded with a DVS due to its high dynamic range. These results indicate the Canny approach
is not suitable for use to train with Pix2Pix.

\section{Conclusion}\label{conclusion}
We demonstrate the feasibility of synthesizing raw data conditionally upon event-driven data. Our exploration of both DVS data as well as synthetic spike-data gives an indication of the adverse effects of high spatial variance, which neuromorphic vision sensors are already suited to mitigate. 
This work can be improved and expanded to time-series data for video reconstruction, and benchmarked using FID to demonstrate not only the quality of image synthesis, but the ability to generate a diverse set of valid reconstructions. Code can be made available upon reasonable request.


 

\end{document}